\documentclass[journal]{IEEEtran}

\usepackage[style=ieee]{biblatex} 
\bibliography{example_bib.bib}    

\usepackage{amsmath}
\usepackage{array}
\usepackage{mdwmath}
\usepackage{mdwtab}
\usepackage{eqparbox}
\usepackage{url}
\usepackage{subfloat}
\usepackage{subfig}

\usepackage{url}
\usepackage{graphicx}  
\usepackage{float}  

\begin{document}

\title{Panoptic Segmentation and Labelling of Lumbar Spine Vertebrae using Modified Attention Unet}

\author{\IEEEauthorblockN{Rikathi Pal\IEEEauthorrefmark{1},
        Priya Saha\IEEEauthorrefmark{2},
        Somoballi Ghoshal\IEEEauthorrefmark{1},
        Amlan Chakrabarti\IEEEauthorrefmark{1}, and
        Susmita Sur-Kolay\IEEEauthorrefmark{3}}\\ 
         \IEEEauthorblockA{
        \IEEEauthorrefmark{1} A. K. Choudhury School of Information Technology, University of Calcutta, Kolkata, India.\\
         \IEEEauthorrefmark{2} Human Physiology, Rammohan College, University of Calcutta, Kolkata. India.\\ 
          \IEEEauthorrefmark{3} ACM Unit, Indian Statistical Institute. Indian Statistical Institute, Kolkata, India.}}
        
        
\markboth{Rikathi Pal}
{Shell \MakeLowercase{\textit{et al.}}: Bare Demo of IEEEtran.cls for IEEE Journals}

\maketitle

\begin{abstract}
Segmentation and labeling of vertebrae in MRI images of the spine are critical for the diagnosis of illnesses and abnormalities. These steps are indispensable as MRI technology provides detailed information about the tissue structure of the spine. Both supervised and unsupervised segmentation methods exist, yet acquiring sufficient data remains challenging for achieving high accuracy. In this study, we propose an enhancing approach based on modified attention U-Net architecture for panoptic segmentation of 3D sliced MRI data of the lumbar spine. Our method achieves an impressive accuracy of 99.5\% by incorporating novel masking logic, thus significantly advancing the state-of-the-art in vertebral segmentation and labeling. This contributes to more precise and reliable diagnosis and treatment planning. \\
\end{abstract}


\section{Introduction}


\IEEEPARstart{W}{ith} the advancement of computer image processing technology, three-dimensional (3D) visualization has emerged as a crucial tool in medical diagnosis, providing professionals with precise and comprehensive information. The applications of three-dimensional (3D) medical picture reconstruction are diverse, ranging from tumor localization to surgical planning. Among the various imaging modalities, T1-weighted and T2-weighted MRI sequences stand out as the most commonly used for non-invasive disease diagnosis \cite{buxton1987contrast}. T1-weighted images are generated using short time to echo (TE) and Short Repetition Time (TR) parameters.

With the development of image processing technology, 3D visualization has become an indispensable method for medical diagnosis \cite{31}, offering doctors a plethora of accurate data. yet there are several difficulties. In \cite{39}, the authors have successfully and effectively reconstructed the 3D of a set of slices along a single axis with 98\% accuracy. In this work, we have used the algorithm for 3D reconstruction and slicing in \cite{39}.

The lumbar spine, comprised of five vertebrae (L1-L5), is pivotal in the human body's structural integrity and functionality. Its significance emanates from its involvement in supporting body weight and facilitating movement. Consequently, precise segmentation and localization of lumbar spine vertebrae emerge as fundamental tasks in medical imaging, contributing profoundly to diagnostics, treatment planning, and therapeutic interventions.
The lumbar spine's crucial role underscores the necessity for accurate segmentation and localization. As the primary weight-bearing region of the spine, any aberration or pathology in the lumbar vertebrae can lead to debilitating conditions such as chronic back pain, nerve compression, or even paralysis. Thus, precise delineation of lumbar vertebrae becomes imperative for clinicians to assess spinal health, identify abnormalities, and formulate appropriate treatment strategies.

Recently, with the rapid development of convolution neural networks (CNN), deep learning-based methods have achieved remarkable performance in semantic segmentation. For medical images, a CNN model trained with complete annotation can obtain accuracy which is at par with that of clinical specialists \cite{briot2018analysis}.
However, manual annotation is laborious and time-consuming which makes it costly, and thus limited annotated data is available within the scope of current literature. Furthermore, due to data sensitive nature of CNN, it can easily fail if it does not have access to different variations during its training procedure. In cases of variations or data obtained from different settings, the model may fail and require retraining on newly obtained data, which thereby increases the overall complexity.

In this work, we have proposed a framework for automatic panoptic segmentation of vertebrae from 3D MRI images reconstructed from a sequence of 2D slices with different slice gaps, using a transformer-based sub-supervised learning approach. The novelties of our method in particular are:
\begin{itemize}
    \item A centroid-based masking technique incorporating features like the area and diameter of each vertebra to generate masks for each 2D slice of MR images 
    \item Modified attention UNET has been used for tumor segmentation incorporating multilabel masking for classifying the vertebra after 3D volume
\end{itemize}

\section{Related Works}
In this section, we discuss briefly the state-of-the-art techniques that are commonly used for image segmentation and we also discuss the existing techniques that are unitized for vertebra segmentation and labeling.

Vertebra segmentation and labeling is a very important problem in the medical imaging domain. It is still a challenge to accurately label vertebrae in human. In CT images, the vertebral hard tissue structure is mostly prominent, many works in labeling vertebra for CT images \cite{xu2023runt}, \cite{qadri2023ct}, \cite{saeed2023automated}, \cite{qadri2023ct}   have been carried out but very few people have worked on MRI spine.  In this section, we discuss about the works that have dealt with this problem either in CT or MRI images.

The authors of  \cite{van2024lumbar} present a comprehensive lumbar spine MRI dataset, publicly available and sourced from multiple centers, encompassing 447 sagittal T1 and T2 MRI series from 218 patients with a history of low back pain, collected across four hospitals. Employing an iterative data annotation method, a segmentation algorithm trained on a subset of the dataset facilitated the semi-automatic segmentation of the remaining images. The algorithm initially provided segmentations, which were reviewed, manually corrected, and incorporated into the training data. Performance benchmarks for both a baseline algorithm and nnU-Net were established, showing comparable results. The evaluation was conducted on a separate set of 39 studies comprising 97 series, additionally facilitating a continuous segmentation challenge for fair algorithm comparison. This initiative aims to foster collaboration in spine segmentation research and enhance the diagnostic utility of lumbar spine MRI.

The study \cite{wang2023deep} developed a deep learning model based on a modified 3D Deeplab V3+ network to automatically segment multiple structures from MR images at the L4/5 level. After data preprocessing, the model underwent five-fold cross-validation, achieving an average Dice Similarity Coefficient (DSC) of 0.886, precision of 0.899, and recall of 0.881 on test sets. Morphometric measurements from 3D reconstruction models generated by manual and automatic segmentation showed no significant differences, indicating the model's accuracy. This automated segmentation approach could enhance lumbar surgical evaluation by facilitating the creation of 3D reconstruction models at the L4/5 level.

With advancements in computer and medical imaging technology\cite{wang2023deep}, medical image segmentation, particularly in magnetic resonance imaging (MRI), has gained prominence in the medical field. MRI offers sensitive detection of tissue changes and physiological information without ionizing radiation, making it valuable for spinal imaging. Manual segmentation by MRI analysts is time-consuming and labor-intensive, necessitating automatic segmentation methods to enhance diagnostic accuracy and aid patient treatment. Artificial intelligence, particularly deep learning, has emerged as a promising approach for automatic segmentation and analysis of spinal MRI scans, facilitating localization, segmentation of spinal structures, diagnosis, differential diagnosis, clinical decision support, and prognosis prediction. Deep learning techniques, such as convolutional neural networks, leverage large datasets to achieve high segmentation accuracy, surpassing 88\% on average, making them effective tools for medical image analysis and processing in spinal imaging.

While significant progress has been made in automating vertebral segmentation and labeling in both CT and MRI images, there are still some drawbacks to current approaches. One limitation lies in the generalization of models across different datasets and imaging conditions. Many algorithms are developed and validated on specific datasets, which may not fully represent the variability encountered in clinical practice. Additionally, the performance of these models can be affected by factors such as image quality, patient characteristics, and anatomical variations. Another challenge is the requirement for large annotated datasets for training deep learning models, which can be time-consuming and resource-intensive to acquire. Furthermore, despite advances in deep learning techniques, these methods may struggle with rare pathologies or cases with ambiguous boundaries, leading to segmentation errors. Addressing these drawbacks will be crucial for enhancing the robustness and applicability of automated vertebral segmentation algorithms in clinical settings.
This paper delves into the mathematical foundations of segmenting and localizing lumbar spine vertebrae, emphasizing the importance of this process for diagnostic precision and patient care. It highlights the challenges in labeling vertebrae in the lumbar region and the advantages of using T2 MRI images. The proposed framework introduces a novel approach for automatically segmenting vertebrae from 3D MRI images reconstructed from a series of 2D slices with varying slice gaps. Two key innovations include a centroid-based masking technique to create masks for each 2D slice and the integration of multilabel masking for vertebra classification after 3D volume reconstruction along the sagittal plane from the sequence of 2D slices. This approach aims to enhance the accuracy and efficiency of lumbar spine analysis, offering potential advancements in spinal health diagnosis and treatment.

\section{Preliminaries}
In this section, we discuss briefly about the existing architecture that has been used as a basis for our proposed method.
\subsection*{\textbf{Attention UNet: }}

The U-Net architecture \cite{ronneberger2015u} is a popular convolutional neural network (CNN) commonly used for semantic segmentation tasks in image processing. It consists of an encoder-decoder structure where the encoder downsamples the input image to extract features, and the decoder upsamples these features to generate a segmentation map.

In the original U-Net architecture, the encoder uses convolutional and pooling layers to extract hierarchical features from the input image. Let $X$ denote the input image, and $Z$ represent the feature maps obtained from the encoder:

\begin{equation}
    Z = E(X)
\end{equation}

where $E$ represents the encoder function.

These features are then propagated through the decoder, which employs transposed convolutions to upsample the features to the original image size while combining them with features from the encoder through skip connections. The skip connections help preserve spatial information and assist in better segmentation accuracy.

However, the standard U-Net architecture lacks mechanisms to selectively focus on relevant regions of the image, which can limit its performance, especially in cases where fine details or subtle features are crucial for accurate segmentation. The Attention U-Net architecture \cite{oktay2018attention} addresses this limitation by integrating attention mechanisms into both the encoder and decoder components.

The attention mechanisms $A_{enc}$ and $A_{dec}$ in the Attention U-Net architecture dynamically compute attention maps that modulate the feature maps extracted by the encoder and decoder, respectively. These attention maps are learned during training and allow the network to focus more on relevant regions while suppressing irrelevant ones. This selective attention mechanism helps improve segmentation accuracy by emphasizing salient features and ignoring distracting or less informative regions.

Mathematically, the attention mechanisms are implemented as learnable functions that take the feature maps $Z$ as input and output attention coefficients. Let $\tilde{Z}_{enc}$ and $\tilde{Z}_{dec}$ denote the modulated feature maps for the encoder and decoder, respectively:

\begin{align}
    \tilde{Z}_{enc} &= A_{enc}(Z) \odot Z \\
    \tilde{Z}_{dec} &= A_{dec}(Z) \odot Z
\end{align}

where $\odot$ denotes element-wise multiplication.

The modulated feature maps are then passed through the decoder to generate the final segmented output:

\begin{equation}
    Y = D(\tilde{Z}_{dec})
\end{equation}

In summary, the Attention U-Net architecture enhances the original U-Net by incorporating attention mechanisms that enable the network to dynamically adapt its focus during the segmentation process, leading to improved accuracy, especially in cases where selective attention to relevant image regions is critical.

\section{Proposed Methodology}

In this study, we provide a novel methodology that uses cutting-edge techniques to improve upon current approaches for the accurate segmentation and labeling of 3D vertebrae. We achieve this by tailoring the attention UNET design to improve its multi-class panoptic segmentation capabilities, which in turn allows for more precise vertebral structure delineation. Taking use of 3D MRI data, we process first by using just sagittal slices that are 5 mm apart. We first convert this data into a 3D model via a painstaking reconstruction procedure, and then we divide it into 2D slices separated by a 1mm interval. We create multi-class labels for vertebral datasets using a unique mask-creation approach. These datasets are used as training data for our newly developed modified attention UNET model. They consist of images and the masks that go with them. Panoptic segmentation of vertebrae is made easier by this thorough process, which holds promise for improvements in diagnostic accuracy and medical imaging analysis.

\subsection{\textbf{Data Preparation and Masking}}
Our proposed method is based on a multi-label panoptic mask using a diameter-based masking procedure. Our dataset consists of T2-weighted MRI. So we prepared a panoptic mask, more specifically, a multi-label panoptic mask that could be used by the model for a feature-rich segmentation.

\begin{figure}
    \centering
        \includegraphics[width=\linewidth, height=2in]{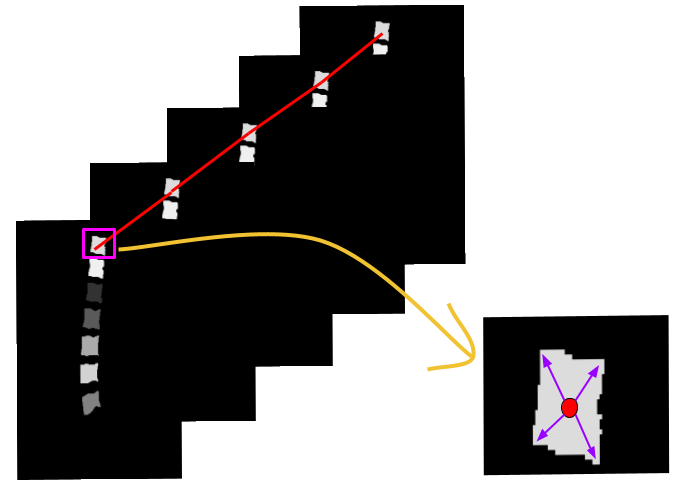}
        \caption{Generating centroid and corner points from reconstructed slices}
        \label{mask}
    \end{figure}

For the required segmentation masks, we first identified and isolated the slices that were properly annotated and had more than one label. The dataset had in total seven labels, five from the lumbar region numbered L1 to L5, and two from the thoracic region numbered T11 and T12. Following this, each of the labels is given a pixel intensity value that ranges from 1 to 7 representing the seven labels respectively. Next, the task is to capture the exact position of a vertebra that is present within the slice. We have information about the centroid and coordinates of two corner points from the centroid for each vertebra per slice, from these, we have calculated the diameter, height-to-width ratio, area, and bounding box of each vertebrae and validated it with \cite{neubert2012automated}. Initially, we have a set of sagittal slices of different patients with a slice gap of $3$ to $5$ mm (1mm = 4 pixels). We have reconstructed a 3D image \cite{39} from the input data and then sliced out the sagittal slices with a 1-pixel slice gap. For the intermediate slices
that we have reconstructed, we don't have the centroid and corner points of each vertebra. We have defined the centroids by taking the orthogonal projection of the centroid points based on the predefined centroids in the previous and next slices as shown in Figure: \ref{mask}. Suppose, we have information of slice number 1 and slice number 15, and we generated the slices between 1 and 15 by using  \cite{39} then, we generate the centroid of each lumbar in slices 2-14 based on the centroid information in slice 1 and 15. Then, based on the diameter information of each vertebra as in  \cite{rasoulian2013lumbar}, we identify the corner points of each vertebra in the generated slices. 
\begin{figure*}
    \centering
    \includegraphics[width=\linewidth, height=1.5in]{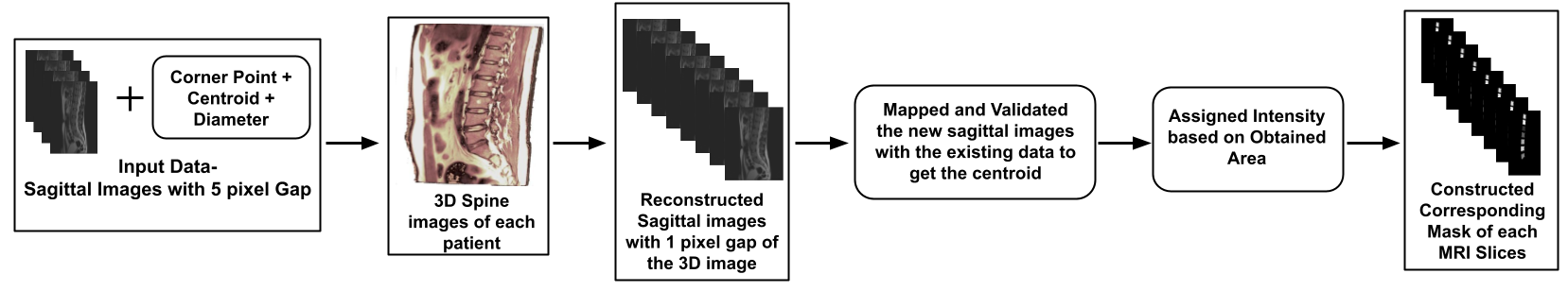}
    \caption{Proposed Masking Process}
    \label{fig:mask}
\end{figure*}

Initially, based on the centroid and two corner coordinates we sketch a bounding box around the vertebrae and then trace the boundary of each vertebra by excluding the “0”
valued or black pixels from the bounding box. We also calculate the diameter by joining the corner points through the centroid. From the corner points we calculate the height-to-width ratio and also the area of each vertebrae. Each vertebrae size and area are validated by comparing our results with \cite{zelditch2012geometric}. The pixels in the obtained bounded region are then changed to the corresponding class intensity value as per the defined label. Figure \ref{fig:mask} shows the step-wise masking process. As the images are inherently grayscale, the individual slices are normalized and cropped. Secondly, while creating the boundary before masking, it is ascertained that for each such step, the boundaries are tightly defined on each side and not a single pixel is more or less than the original size of the vertebra in the MRI scan.

\subsection{\textbf{Model Training for Panoptic Segmentation}}
\begin{figure*}
	\centering
\subfloat[]{\includegraphics[width=\linewidth, height=2in]{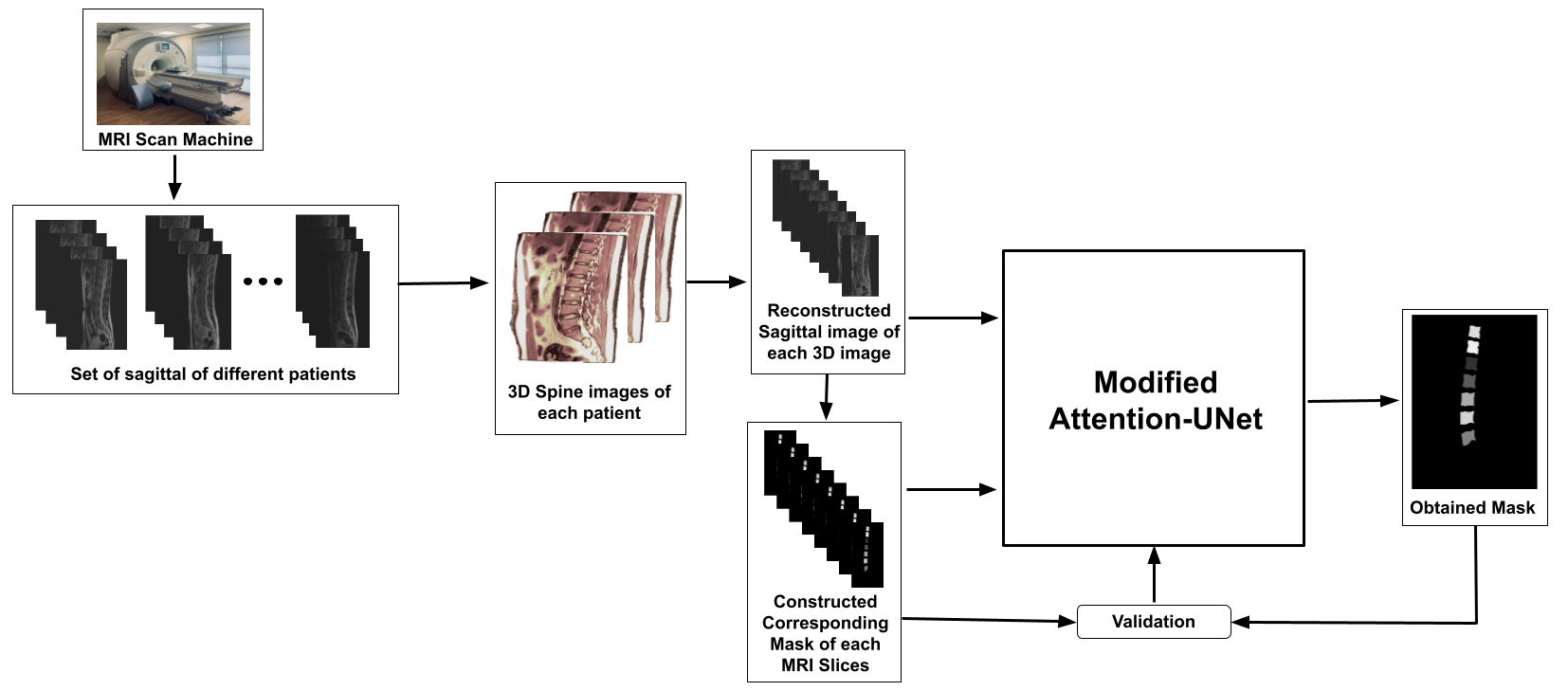}}
	\hspace{.02 mm}
	\subfloat[]	{\includegraphics[width=\linewidth, height=2in]{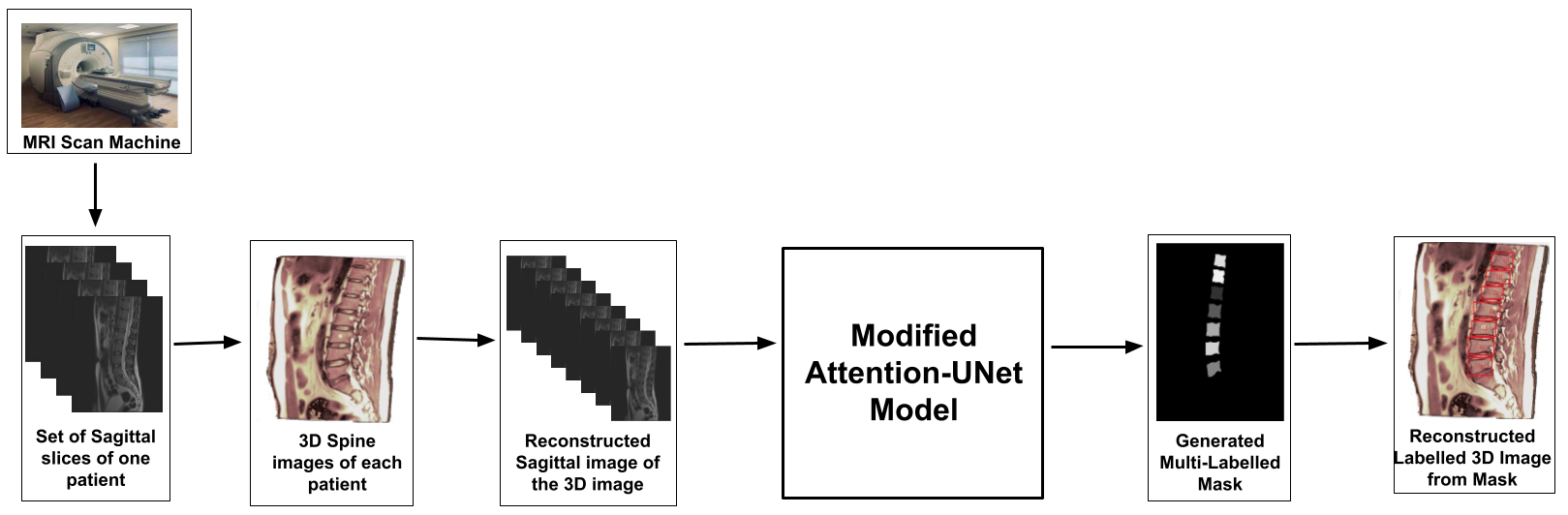}}
	     \caption{Block diagram of the proposed method (a) Training Phase (b) Testing Phase }	
     \label{3Dspine}
\vspace{0.01 mm}
\end{figure*}

The comprehensive workflow showing the training and testing phases of the Modified version of the Attention UNet model for Panoptic segmentation is depicted in Figure \ref{3Dspine}.

\subsubsection{Input Preprocessing: }
The input images undergo preprocessing to optimize them for subsequent network processing. This preprocessing involves two main steps: denoising and resizing. Firstly, the images are subjected to denoising techniques like shearlet transform \cite{24} to reduce any unwanted noise or artifacts present in the raw data. This ensures that the input data is clean and free from unnecessary distractions, thereby enhancing the network's ability to focus on relevant features during training and inference. Following denoising, the images are reshaped to a standardized size of 256$\times$256 pixels. Resizing the images to a consistent dimension facilitates uniformity in input data across the dataset, which is crucial for achieving consistent performance and generalization of the neural network model. By pre-processing the input images in this manner, we prepare them effectively for further processing by the network, ultimately improving the overall performance and robustness of the system.

\subsubsection{Proposed Modified Attention UNet Architecture: }
In the context of semantic segmentation, the Attention UNet model, denoted as \( M \), serves as a fundamental architecture. It typically comprises four encoder (\( L_e \)) and four decoder (\( L_d \)) layers, resulting in a total of \( L = L_e + L_d = 8 \) layers. However, in our proposed modified version, denoted as \( M' \), we introduce an extended architecture with six layers in both the encoder and decoder sections, yielding \( L = L_e + L_d = 12 \) layers. This expanded depth facilitates more comprehensive feature extraction and resolution recovery, essential for accurate panoptic segmentation which includes the output instance IDs for each pixel belonging to individual objects in the scene.

To address the challenge of accuracy degradation with increased layer depth, we introduce the downsampling factor \( f \) to quantify the loss of spatial information during downsampling. We define the function \( \phi(f) \) to maintain a balance between depth and resolution preservation:

\[ \phi(f) = \text{Balance between depth and resolution preservation} \]

Through experimentation, we observe that simply increasing the number of layers (\( \Delta L \)) may not proportionally improve performance (\( \Delta M \)), as downsampling can lead to the loss of crucial spatial information. Hence, our model is carefully designed to ensure that deeper layers capture intricate features while preserving spatial details vital for accurate segmentation.

The enhanced performance of the modified Attention UNet (\( M' \)) in panoptic segmentation tasks is quantified by the improvement in both feature representation and spatial resolution recovery:
\begin{equation}
\resizebox{0.91\hsize}{!}{%
 $\text{Improvement}(M') = \text{Enhancement}(M')_{\text{feature}} + \text{Enhancement}(M')_{\text{resolution}}$}
\end{equation}

This strategic integration of additional layers within the UNet architecture enables our model to outperform standard counterparts, effectively enhancing both feature representation and spatial resolution recovery.

\begin{figure*}[h]

    \centering
        \includegraphics[width=\linewidth, height=2in]{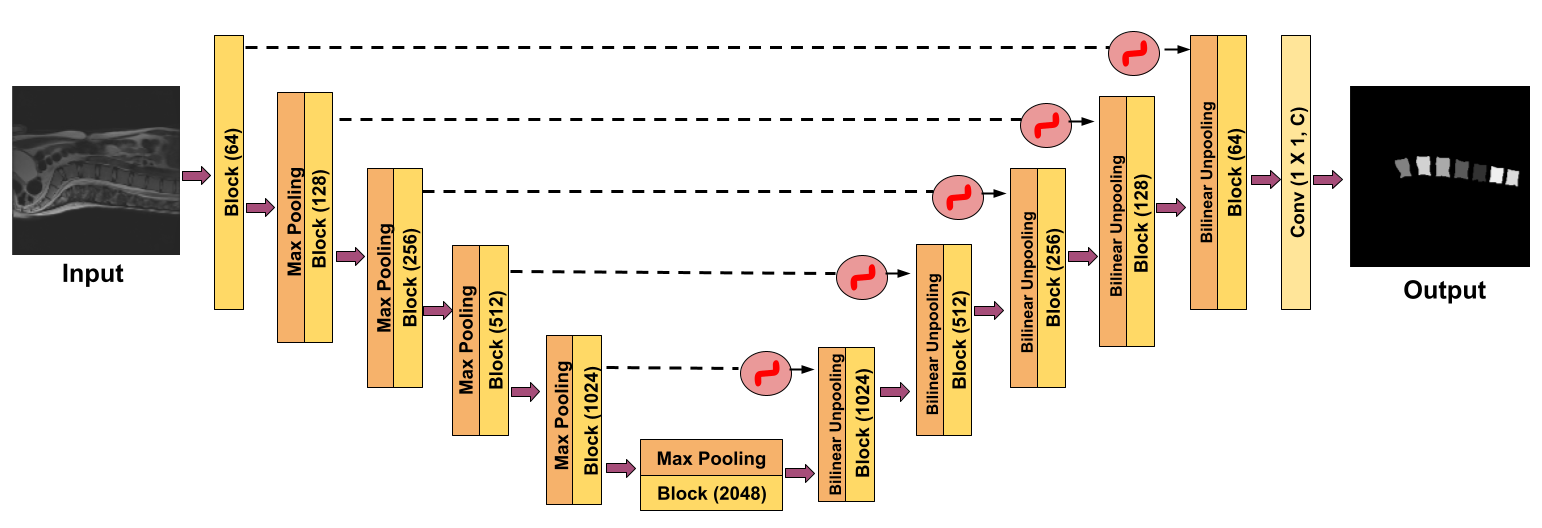}
        \caption{Attention Unet Model}
        \label{model}
    \end{figure*}

The Modified Attention UNet architecture as shown in figure: \ref{model} is constructed by assembling six layers of the encoder, attention mechanism, and decoder components into a unified deep learning model. The model takes input images and produces corresponding segmentation masks in an end-to-end manner, leveraging both local and global contextual information through the attention mechanism.

\subsubsection{Training and Evaluation: }
The proposed modified attention UNet model undergoes training utilizing suitable optimization algorithms and loss functions, such as binary cross-entropy. These mechanisms aim to minimize the disparity between predicted segmentation masks and ground truth annotations. During training, a combination of both semantic segmentation loss i.e., cross-entropy loss, and instance segmentation loss i.e. instance-aware segmentation loss, is employed. This dual-loss strategy ensures that the model optimizes both for accurate semantic segmentation, discerning different classes within an image, and instance segmentation, distinguishing individual instances of the same class.
After training, the performance evaluation of the model is conducted using standard metrics. These metrics encompass visualizations like the Loss to Epoch Graph, which illustrates the convergence of the model's loss function over training epochs. Additionally, metrics such as Intersection over Union (IoU), measuring the overlap between predicted and ground truth segmentation masks, and accuracy, gauging the overall correctness of pixel-wise predictions, are employed.
In post-processing, following inference, additional steps are often taken to refine the model's output. This entails applying techniques to consolidate and improve the segmentation results. For instance, post-processing may involve grouping pixels with the same semantic label and instance ID into coherent object masks. These refinements contribute to enhancing the overall quality and coherence of the segmentation outputs generated by the model.

\subsubsection{Output Generation: }
In the process of the decoding, the final decoder block orchestrates the generation of the output segmentation mask. This pivotal step involves the utilization of a convolution layer with sigmoid activation, a technique instrumental in yielding pixel-wise predictions. Through this intricate process, the network synthesizes its comprehension of each pixel within the input images, encapsulating its discernment of semantic boundaries and instance delineations. As a result, the output segmentation mask serves as a tangible manifestation of the model's interpretative prowess, embodying its capacity to discern and categorize visual elements with precision.
    
\subsubsection{Testing: }
In the testing phase, a series of sagittal slices from each patient's dataset is collected. These slices serve as the foundational input for the subsequent 3D spine reconstruction process. Through meticulous reconstruction, the sagittal slices are arranged with a precise 1 mm gap, ensuring comprehensive coverage and continuity.

Each of these meticulously arranged slices is then fed into the trained Attention UNet Model. Leveraging the insights gleaned from its training, the modified Attention UNet Model embarks on the task of generating multi-class masks. These masks, representative of various anatomical structures and pathological features, form the basis for the ensuing construction of the 3D lumbar spine image.

Following the generation of the predicted masks, a crucial evaluation ensues. The predicted masks are juxtaposed against their corresponding actual masks, facilitating a comprehensive comparison. This meticulous analysis serves as the linchpin for assessing the accuracy and efficacy of the model's predictions. Through this rigorous process of evaluation, the model's performance and reliability in delineating lumbar spine structures are meticulously scrutinized and validated.

\begin{figure}[]
\centering
\subfloat[]{\includegraphics[width=3cm, height=2.5cm]{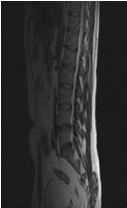}}
\hspace{1cm}
\subfloat[]{\includegraphics[width=3cm, height=2.5cm]{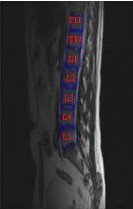}}
\hspace{1cm}
	\caption{Example 1 for Vertebra labeling in 2D lumbar sagittal slice -- (a) Input image, (b) Output labeled vertebra}	
	\label{r1}
\end{figure}

\section{Results}

\begin{figure}[!h]
\centering
\subfloat[]{\includegraphics[width=3cm, height=2.5cm]{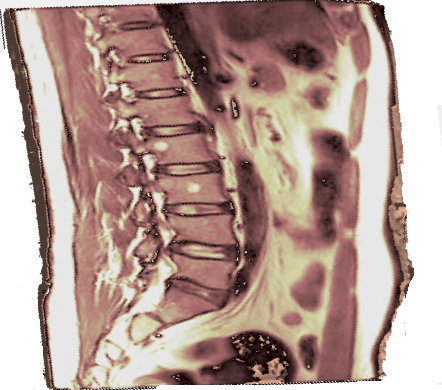}}
\hspace{0.5cm}
\subfloat[]{\includegraphics[width=3cm, height=2.5cm]{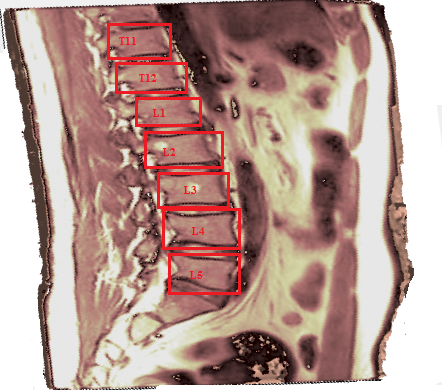}}
\hspace{0.5cm}
\caption{Example of vertebra labeling in 3D MRI of spine -- \\(a) 3D reconstructed image from slices, and (b) vertebra labeled 3D image}
\label{fig:example3dlabel}
\end{figure}

In this section, we briefly discuss the results generated from our proposed method. Figure: \ref{r1} shows an example of 2D and Figure: \ref{fig:example3dlabel} shows me an example of a 3D spine image.

\subsection{\textbf{Dataset Description}}

We have applied our algorithm on PACS 24 T2 MRI lumbar spine labeled data set \cite{pacslumbar}, eight T2 MRI lumbar spine \cite{spineweb}, and on twenty-three real-life T2 MRI data of lumbar spine collected from Bangur Institute of Neurosciences, Kolkata. Our proposed model has been trained with $80\%$ of the dataset, and the rest $20\%$ is used for testing. For training and testing, the images were randomly chosen and fed to our model. For testing, we obtained the 3D volumetric image set by combining the 2D slices along the sagittal plane. The model is then used to predict the mask on a per-slice basis along the sagittal axis, thereby allowing us to obtain the full volumetric mask.

\subsection{\textbf{Experimental Configuration and Parameterization}}
Meticulous performance analysis and comparative study were carried out on a computing platform featuring a high-performing Intel i5 processor with a clock speed of 4.90 GHz and 16 gigabytes of RAM. The GPU of NVIDIA Tesla K80 with 12GB of VRAM, and the GPU runtime includes an Intel Xeon CPU @2.20 GHz, 13 GB RAM, and 12 GB GDDR5 VRAM is used for training purposes. The initial training phase involved a judicious selection of a mere 0.05\% of the remaining unlabelled data, laying the foundation for subsequent analyses. The crucial parameter $\alpha$ was set to 0.6, influencing the model's behavior throughout the training and testing phases. Significantly, distinct $k$-neighborhood values were employed for individual datasets, allowing for a nuanced examination of their impact. This thorough approach not only showcases the effectiveness of the Intel i5 processor and RAM allocation but also provides valuable insights into the intricate interplay of parameters influencing the overall system performance.

\subsection{\textbf{Performance Evaluation Metrics}}

We have chosen three separate metrics during the training process. For the prediction of correct masking topology, Intersection-over-Union (IoU) \cite{rezatofighi2019generalized} has been chosen. As we are also dealing with a 3D multi-class classification problem, we have taken class-based accuracy. Finally, there was logit loss which is used to measure the overall performance of the model for per-pixel classification.

In our case, if an image contains a vertebra correctly identified, it counts as a true positive ($TP$). If a vertebra is present in the image but not identified at all, it counts as a false negative ($FN$). If another vertebra or some other tissue is falsely identified in place of a vertebra, it counts as a false positive ($FP$). If there is no vertebra in the image and it is not identified, it counts as a true negative ($TN$). The Intersection over Union (IoU) is expressed as:
\begin{equation}
IoU = \frac{TP}{TP + FP + FN}
\end{equation}
It is computed as the mean over all the training samples in a batch. During training, the step mean IoU is calculated over a range, providing feedback to the model during a single step and correcting it over the next batch.

The 3D multi-class accuracy \cite{smith2003effects} is given by:
\begin{equation}
{Class Accuracy} = \frac{TN + TP}{TN + TP + FP + FN}
\end{equation}

We also consider the loss of the model over its training process, represented as an average loss in Figure.~\ref{loss}. The model is trained with native pixel-to-pixel cross-entropy loss \cite{NEURIPS2018_f2925f97}. Cross entropy is defined as:
\begin{equation}
L_{CE} = -\sum^{n}_{i=1} t_{i} \log(p_{i})
\end{equation}
Here, $t_{i}$ represents the true class value of each labeled pixel, and $p_{i}$ is the Softmax probability of the $i$-th class predicted by the model.

The Dice score \cite{bertels2019optimizing} is a common metric used in image segmentation tasks to evaluate the similarity between two binary images. It is particularly useful when dealing with imbalanced datasets, where the number of foreground (positive) pixels vastly differs from the number of background (negative) pixels.

The Dice score, also known as the Dice coefficient or Dice similarity coefficient, is defined as:

\begin{equation}
\text{Dice} = \frac{2 \times |X \cap Y|}{|X| + |Y|}
\end{equation}

where:
\begin{itemize}
    \item $X$ and $Y$ are the sets of pixels in the predicted segmentation mask and the ground truth mask, respectively.
    \item $|X \cap Y|$ denotes the number of pixels where both the predicted and ground truth masks have positive values (i.e., true positives).
    \item $|X|$ and $|Y|$ represent the total number of positive pixels in the predicted and ground truth masks, respectively.
\end{itemize}

The Dice score ranges from 0 to 1, where a score of 1 indicates a perfect overlap between the predicted and ground truth masks, while a score of 0 indicates no overlap at all.

\subsection{\textbf{Comparative Study}}
\begin{figure}[h]

    \centering
        \includegraphics[width=\linewidth, height = 2in]{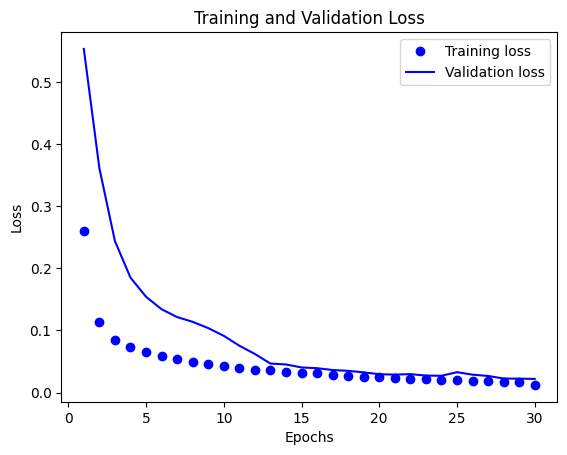}
        \caption{Loss during training phase}
        \label{loss}
    \end{figure}

\begin{figure}[h]

    \centering
        \includegraphics[width=\linewidth, height = 2in]{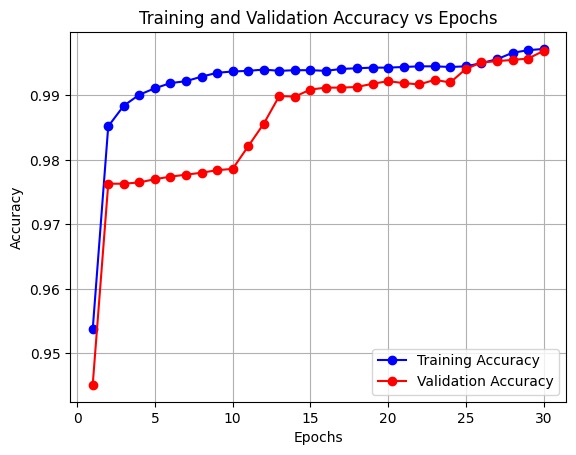}
        \caption{Accuracy during Training Phase}
        \label{Accuracy}
    \end{figure}

In examining the loss curve illustrated in Figure \ref{loss}, a notable trend emerges wherein both training and validation losses exhibit a steady decline over time. Simultaneously, the accuracy curve depicted in Figure \ref{Accuracy} exhibits a consistent upward trajectory during the training process. This coherent pattern underscores the strong performance of our model.

\begin{figure}[h]

    \centering
        \includegraphics[width=\linewidth, height=2in]{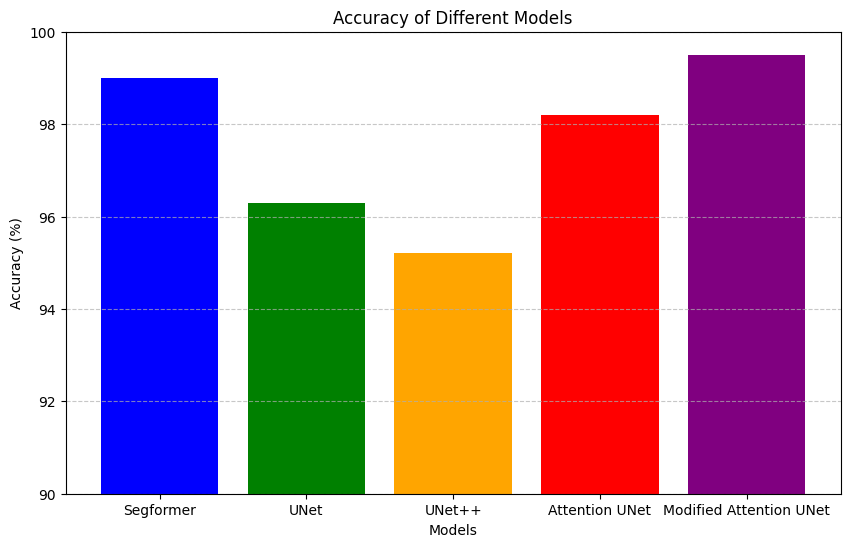}
        \caption{Accuracy Comparison}
        \label{Histogram}
    \end{figure}

Furthermore, a comparative analysis was conducted between our modified Attention UNet Architecture and four state-of-the-art (SOTA) techniques, as presented in Figure \ref{Histogram}. Encouragingly, our model outperforms the existing methodologies with 99.5\% accuracy, showcasing its superiority in delivering favorable results.

\begin{figure}[h]

    \centering
        \includegraphics[width=\linewidth, height = 2in]{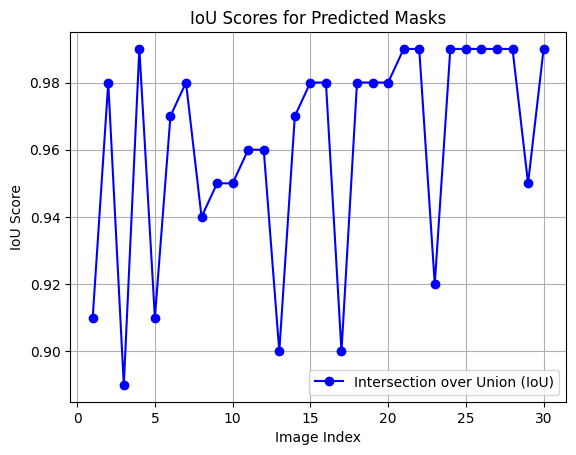}
        \caption{Intersection Over Union during Testing Phase}
        \label{IOU}
    \end{figure}

In addition to performance metrics like loss and accuracy, we also assessed the quality of masks using Intersection over Union (IOU) Scores, as shown in Figure \ref{IOU}. Remarkably, all IOU scores associated with the predicted masks surpass the 0.9 threshold, indicating the robustness of our model across various images. This comprehensive evaluation underscores the efficacy and reliability of our approach in accurately delineating spine tumors. Such promising results underscore the potential of our methodology for clinical application and diagnostic purposes.

We have compared the work with various existing methods as shown in Table: \ref{tab:compare} and we can see that our model outperformed the existing methods.

\begin{table}[h]
 \caption{Comparison with earlier Segmentation Methods}
  \centering
\label{tab:compare}
\begin{tabular}{|c|c|c|}
\hline
 Publications & Accuracy & Dice Score \\
 \hline
Huang et al. \cite{huang2009learning} & 96.70\% & 0.95 \\
\hline
Isensee et al. \cite{isensee2021nnu} & 98.60\% & 0.96 \\
\hline
Wang et al. \cite{wang2023deep} & 93\% & 0.93 \\
\hline
Rasoulian et al. \cite{rasoulian2013lumbar} & 83.77\% & 0.81 \\
\hline
Sekuboyina et al. \cite{sekuboyina2021verse} & 92.70\% & 0.89 \\
\hline
Juang et al. \cite{juang2010mri} & 96\% & 0.94\\
\hline
Zhang et al. \cite{zhang2021spine} & 94.54\% & 0.92 \\
\hline
\textbf{Modified Attention U-Net} & \textbf{99.70\%} & \textbf{0.98}\\
\hline
\end{tabular}
\end{table}

\section{Conclusion and Future Works}
Through the automated labeling of lumbar spine vertebrae, our proposed research seeks to advance the field of medical imaging. Using statistical computations of mask dimensions from a small sample of labeled data in T2 MRI lumbar spine images, we can reliably mark the correct vertebrae labels and reconstruct 3D pictures with minimal gaps. With a Modified Attention UNet architecture, we attain a remarkable 99.5\% tagging accuracy. Especially, our new mask generation method using geometric and statistical models effectively tackles data scarcity problems and supports segmentation procedures. Moreover, the adaptability of our technology is further improved by our multi-class vertebral labeling process. Our methodology exceeds existing methods with an automatic vertebral tagging accuracy of 99.5\%, offering significant improvements in customized medicine. Furthermore, our method's prospective integration with robotic surgical instruments has the potential to transform surgical planning and enable more accurate and effective surgeries. Our ultimate goal is to provide medical practitioners with state-of-the-art tools for better diagnosis and treatment planning, which will ultimately result in better patient outcomes.

\section*{Acknowledgment}
We extend our heartfelt gratitude to the esteemed professionals whose invaluable contributions have enriched our understanding of complex data:
Dr. Alok Pandit, from the Bangur Institute of Neurosciences, IPGMER Kolkata, whose expertise and guidance have illuminated intricate aspects of our analysis.
Their unwavering commitment to advancing knowledge and their willingness to share their expertise have been instrumental in our pursuit of understanding. We are deeply grateful for their generous support and guidance.

\printbibliography

@article{xu2023runt,
  title={RUnT: A network combining residual U-Net and transformer for vertebral edge feature fusion constrained spine CT image segmentation},
  author={Xu, Hao and Cui, Xinxin and Li, Chaofan and Tian, Zhenyu and Liu, Jing and Yang, Jianlan},
  journal={IEEE Access},
  year={2023},
  publisher={IEEE}
}

@article{qadri2023ct,
  title={CT-based automatic spine segmentation using patch-based deep learning},
  author={Qadri, Syed Furqan and Lin, Hongxiang and Shen, Linlin and Ahmad, Mubashir and Qadri, Salman and Khan, Salabat and Khan, Maqbool and Zareen, Syeda Shamaila and Akbar, Muhammad Azeem and Bin Heyat, Md Belal and others},
  journal={International Journal of Intelligent Systems},
  volume={2023},
  pages={1--14},
  year={2023},
  publisher={Hindawi Limited}
}

@article{saeed2023automated,
  title={An automated deep learning approach for spine segmentation and vertebrae recognition using computed tomography images},
  author={Saeed, Muhammad Usman and Dikaios, Nikolaos and Dastgir, Aqsa and Ali, Ghulam and Hamid, Muhammad and Hajjej, Fahima},
  journal={Diagnostics},
  volume={13},
  number={16},
  pages={2658},
  year={2023},
  publisher={MDPI}
}

@article{van2024lumbar,
  title={Lumbar spine segmentation in MR images: a dataset and a public benchmark},
  author={van der Graaf, Jasper W and van Hooff, Miranda L and Buckens, Constantinus FM and Rutten, Matthieu and van Susante, Job LC and Kroeze, Robert Jan and de Kleuver, Marinus and van Ginneken, Bram and Lessmann, Nikolas},
  journal={Scientific Data},
  volume={11},
  number={1},
  pages={264},
  year={2024},
  publisher={Nature Publishing Group UK London}
}

@article{wang2023deep,
  title={Deep Learning-Based Automated Magnetic Resonance Image Segmentation of the Lumbar Structure and Its Adjacent Structures at the L4/5 Level},
  author={Wang, Min and Su, Zhihai and Liu, Zheng and Chen, Tao and Cui, Zhifei and Li, Shaolin and Pang, Shumao and Lu, Hai},
  journal={Bioengineering},
  volume={10},
  number={8},
  pages={963},
  year={2023},
  publisher={MDPI}
}

@article{buxton1987contrast,
  title={Contrast in rapid MR imaging: T1-and T2-weighted imaging},
  author={Buxton, Richard B and Edelman, Robert R and Rosen, Bruce R and Wismer, Gary L and Brady, Thomas J},
  journal={J Comput Assist Tomogr},
  volume={11},
  number={1},
  pages={7--16},
  year={1987}
}

@inproceedings{briot2018analysis,
  title={Analysis of efficient cnn design techniques for semantic segmentation},
  author={Briot, Alexandre and Viswanath, Prashanth and Yogamani, Senthil},
  booktitle={Proceedings of the IEEE Conference on Computer Vision and Pattern Recognition Workshops},
  pages={663--672},
  year={2018}
}

@inproceedings{ronneberger2015u,
  title={U-net: Convolutional networks for biomedical image segmentation},
  author={Ronneberger, Olaf and Fischer, Philipp and Brox, Thomas},
  booktitle={Medical image computing and computer-assisted intervention--MICCAI 2015: 18th international conference, Munich, Germany, October 5-9, 2015, proceedings, part III 18},
  pages={234--241},
  year={2015},
  organization={Springer}
}

@article{oktay2018attention,
  title={Attention u-net: Learning where to look for the pancreas},
  author={Oktay, Ozan and Schlemper, Jo and Folgoc, Loic Le and Lee, Matthew and Heinrich, Mattias and Misawa, Kazunari and Mori, Kensaku and McDonagh, Steven and Hammerla, Nils Y and Kainz, Bernhard and others},
  journal={arXiv preprint arXiv:1804.03999},
  year={2018}
}

@article{31,
author = {Herghelegiu, P. and Gavrilescu, M. and Manta, Vasile},
year = {2011},
month = {08},
pages = {99-104},
title = {Visualization of Segmented Structures in 3D Multimodal Medical Data Sets},
volume = {11},
journal = {Advances in Electrical and Computer Engineering},
doi = {10.4316/aece.2011.03016}
}

@article{39,
author = {Ghoshal, Somoballi and Banu, Sourav and Chakrabarti, Amlan and Sur Kolay, Susmita and Pandit, Alok},
year = {2020},
month = {10},
pages = {},
title = {3D Reconstruction of Spine Image from 2DMRI Slices along One Axis},
volume = {14},
journal = {IET Image Processing},
doi = {10.1049/iet-ipr.2019.0800}
}

@article{24,
author = {Shahdoosti, Hamid and Khayat, Omid},
year = {2016},
month = {09},
pages = {},
title = {Image denoising using sparse representation classification and non-subsampled shearlet transform},
volume = {10},
journal = {Signal, Image and Video Processing},
doi = {10.1007/s11760-016-0862-0}
}

@misc{pacslumbar,
title = {\url{http://dx.doi.org/10.5281/zenodo.22304}},
  note = {Accessed: 2023-03-07}}

@misc{spineweb,
title = {\url{http://spineweb.digitalimaginggroup.ca/}},
  note = {Accessed: 2017-03-01}
}

@article{huang2009learning,
  title={Learning-based vertebra detection and iterative normalized-cut segmentation for spinal MRI},
  author={Huang, Szu-Hao and Chu, Yi-Hong and Lai, Shang-Hong and Novak, Carol L},
  journal={IEEE transactions on medical imaging},
  volume={28},
  number={10},
  pages={1595--1605},
  year={2009},
  publisher={IEEE}
}

@article{isensee2021nnu,
  title={nnU-Net: a self-configuring method for deep learning-based biomedical image segmentation},
  author={Isensee, Fabian and Jaeger, Paul F and Kohl, Simon AA and Petersen, Jens and Maier-Hein, Klaus H},
  journal={Nature methods},
  volume={18},
  number={2},
  pages={203--211},
  year={2021},
  publisher={Nature Publishing Group}
}

@article{sekuboyina2021verse,
  title={VerSe: A Vertebrae labelling and segmentation benchmark for multi-detector CT images},
  author={Sekuboyina, Anjany and Husseini, Malek E and Bayat, Amirhossein and L{\"o}ffler, Maximilian and Liebl, Hans and Li, Hongwei and Tetteh, Giles and Kuka{\v{c}}ka, Jan and Payer, Christian and {\v{S}}tern, Darko and others},
  journal={Medical image analysis},
  volume={73},
  pages={102166},
  year={2021},
  publisher={Elsevier}
}

@article{rasoulian2013lumbar,
  title={Lumbar spine segmentation using a statistical multi-vertebrae anatomical shape+ pose model},
  author={Rasoulian, Abtin and Rohling, Robert and Abolmaesumi, Purang},
  journal={IEEE transactions on medical imaging},
  volume={32},
  number={10},
  pages={1890--1900},
  year={2013},
  publisher={IEEE}
}

@article{zhang2021spine,
  title={Spine medical image segmentation based on deep learning},
  author={Zhang, Qingfeng and Du, Yun and Wei, Zhiqiang and Liu, Hengping and Yang, Xiaoxia and Zhao, Dongfang and others},
  journal={Journal of Healthcare Engineering},
  volume={2021},
  year={2021},
  publisher={Hindawi}
}

@inproceedings{rezatofighi2019generalized,
  title={Generalized intersection over union: A metric and a loss for bounding box regression},
  author={Rezatofighi, Hamid and Tsoi, Nathan and Gwak, JunYoung and Sadeghian, Amir and Reid, Ian and Savarese, Silvio},
  booktitle={Proceedings of the IEEE/CVF conference on computer vision and pattern recognition},
  pages={658--666},
  year={2019}
}

@inproceedings{bertels2019optimizing,
  title={Optimizing the dice score and jaccard index for medical image segmentation: Theory and practice},
  author={Bertels, Jeroen and Eelbode, Tom and Berman, Maxim and Vandermeulen, Dirk and Maes, Frederik and Bisschops, Raf and Blaschko, Matthew B},
  booktitle={Medical Image Computing and Computer Assisted Intervention--MICCAI 2019: 22nd International Conference, Shenzhen, China, October 13--17, 2019, Proceedings, Part II 22},
  pages={92--100},
  year={2019},
  organization={Springer}
}

@article{smith2003effects,
  title={Effects of landscape characteristics on land-cover class accuracy},
  author={Smith, Jonathan H and Stehman, Stephen V and Wickham, James D and Yang, Limin},
  journal={Remote Sensing of Environment},
  volume={84},
  number={3},
  pages={342--349},
  year={2003},
  publisher={Elsevier}
}

@book{zelditch2012geometric,
  title={Geometric morphometrics for biologists: a primer},
  author={Zelditch, Miriam and Swiderski, Donald and Sheets, H David},
  year={2012},
  publisher={academic press}
}

@article{neubert2012automated,
  title={Automated detection, 3D segmentation and analysis of high resolution spine MR images using statistical shape models},
  author={Neubert, Ale{\v{s}} and Fripp, Jurgen and Engstrom, Craig and Schwarz, Raphael and Lauer, Lars and Salvado, Olivier and Crozier, Stuart},
  journal={Physics in Medicine \& Biology},
  volume={57},
  number={24},
  pages={8357},
  year={2012},
  publisher={IOP Publishing}
}

@article{juang2010mri,
  title={MRI brain lesion image detection based on color-converted K-means clustering segmentation},
  author={Juang, Li-Hong and Wu, Ming-Ni},
  journal={Measurement},
  volume={43},
  number={7},
  pages={941--949},
  year={2010},
  publisher={Elsevier}
}

@inproceedings{NEURIPS2018_f2925f97,
 author = {Zhang, Zhilu and Sabuncu, Mert},
 booktitle = {Advances in Neural Information Processing Systems},
 editor = {S. Bengio and H. Wallach and H. Larochelle and K. Grauman and N. Cesa-Bianchi and R. Garnett},
 pages = {},
 publisher = {Curran Associates, Inc.},
 title = {Generalized Cross Entropy Loss for Training Deep Neural Networks with Noisy Labels},
 volume = {31},
 year = {2018}
}

\end{document}